\pdfoutput=1

\documentclass[11pt]{article}

\usepackage[]{ACL2023}

\usepackage{amsmath,amsfonts,bm}
\usepackage{bbm}
\usepackage{tikz}
\usetikzlibrary{plotmarks}

\usepackage{amsmath,amsfonts,bm}

\def\eqref#1{equation~\ref{#1}}

\def\1{\bm{1}}

\DeclareMathAlphabet{\mathsfit}{\encodingdefault}{\sfdefault}{m}{sl}
\SetMathAlphabet{\mathsfit}{bold}{\encodingdefault}{\sfdefault}{bx}{n}
\usepackage{times}
\usepackage{latexsym}

\usepackage[T1]{fontenc}

\usepackage[utf8]{inputenc}

\usepackage{microtype}

\usepackage{inconsolata}

\usepackage{microtype}
\usepackage{hyperref}
\usepackage{url}
\usepackage{booktabs}
\definecolor{darkblue}{rgb}{0, 0, 0.5}
\hypersetup{colorlinks=true, citecolor=darkblue, linkcolor=darkblue, urlcolor=darkblue}

\usepackage{amssymb}
\usepackage{amsfonts}
\usepackage{mathrsfs}
\usepackage{dsfont}
\usepackage{booktabs}
\usepackage{graphicx}
\usepackage{multirow}
\usepackage{arydshln}
\usepackage{booktabs} 
\usepackage{caption}
\usepackage{xspace}
\usepackage{enumitem}
\usepackage{tcolorbox}
\usepackage{soul}
\usepackage{wrapfig,lipsum,booktabs}
\usepackage{float}

\newcommand{\supervised}{Src+Tgt\xspace}
\newcommand{\mlm}{CPT\xspace}
\newcommand{\mmd}{UDA\footnotesize PTER\normalsize\xspace}

\title{How Useful is Continued Pre-Training for Generative Unsupervised Domain Adaptation?}

\author{
\begin{tabular}{ccc}
    Rheeya Uppaal$^1$ & Yixuan Li$^1$ & Junjie Hu$^{1,2}$
\end{tabular}
\\
  $^1$Department of Computer Sciences, \\$^2$Department of Biostatistics and Medical Informatics \\
  University of Wisconsin-Madison \\
  \texttt{\{uppaal, jhu, sharonli\}@cs.wisc.edu} \\
  }

\begin{document}
\maketitle

\begin{abstract}
    Recent breakthroughs in scale have enabled the emergence of powerful generative language models, and the ability to fine-tune these models on various tasks by casting them into prompts or instructions.
    In this landscape, the problem of Unsupervised Domain Adaptation (UDA), or the problem of leveraging knowledge from a labeled source domain to an unlabeled target domain, has been left behind, with recent UDA methods still addressing discriminative classification. In particular, two popular UDA approaches, involving Continued Pre-Training (\mlm) and learning domain invariant representations, have been under-explored in the generative setting, signaling a gap.
    In this work, we evaluate the utility of \mlm for generative UDA. 
    We first perform an empirical evaluation to measure the trade-offs between~\mlm and strong methods promoting domain invariance. 
    We further evaluate how well the benefits of~\mlm extend to different architectures, tuning methods and data regimes.
    We then motivate the use of~\mlm by studying to what degree it benefits classification performance on the target domain.
    Finally, we attempt to understand the mechanism behind which~\mlm improves classification performance on the unlabeled target domain.
    Our findings suggest that the model implicitly learns the downstream task while predicting masked words informative to that task.
    Our work connects the body of UDA research with that of instruction tuning, enabling an initial step towards a wider applicability of modern language models.
    Our code is available at \url{https://github.com/Uppaal/cpt-generative-uda}.
\end{abstract}

\section{Introduction}
\label{sec:introduction}

Recent advancements in the pre-training of language models have enabled the widespread use of powerful generative models, which can be leveraged across multiple domains with no training~\citep[\textit{inter alia}]{brown2020language, scao2022bloom, touvron2023llama}. 
Despite these advancements, these autoregressive models are still fragile under certain kinds of data distribution shifts, making their applications across these domains challenging~\citep[\textit{inter alia}]{ribeiro2020beyond, bajaj2021long, chuang2023evolving, uppaal2024detox}. 
This is addressed, in part, by the concept of instruction tuning with templates~\citep{zhang2023instruction, sanhmultitask, ouyang2022training, wang2022super, weifinetuned}, enabling the learning of new tasks without any randomly initialized parameters.

The problem of unsupervised domain adaptation (UDA) leverages learned knowledge from a labeled source domain to an unlabeled target domain~\citep[\textit{inter alia}]{pan2010survey, ganin2015unsupervised, long2015learning}. 
It is useful for adaptation to unlabeled domains with high labeling costs, where supervised instruction tuning does not suffice.
Despite the pervasive need for models to generalize to such domains, recent UDA methods still address discriminative classification, 
barring the application of these approaches to recent generative models. 
In particular, Continued Pre-Training and Domain Invariance-based methods, two widely popular classes of UDA approaches~\citep{ramponi2020neural}, are completely unexplored for UDA in the generative setting.

The Continued Pre-Training (\mlm) approach involves extended pre-training on a domain or task, followed by supervised training on the downstream task~\citep{gururangan2020don}. This approach has been widely used for adaptation to  labeled domains~\citep[\textit{inter alia}]{gao2021making, kim2021revisiting, hung2023tada}, and in the UDA setup for unlabeled domains~\citep{han2019unsupervised, zhangunsupervised, karouzos2021udalm, pfeiffer2020mad, parovic2023cross}. Invariance-based approaches attempt to learn representations that are invariant across domains~\citep{tzeng2014deep, ganin2016domain, wu2022adversarial, guo-etal-2022-improving}, with the notion that when the learned representations from both domains cannot be distinguished by a classifier and the classifier performs well on the source domain, it will also exhibit strong performance on the target domain.
These two classes of methods introduce a trade-off: invariance-based methods suffer from instability issues~\citep{han2019unsupervised, ramesh2021domain}, while continued pre-training requires a larger computational budget. But how would this trade-off extrapolate to the generative setting? For example, invariance-based methods are well motivated in discriminative tasks, where there is a clear decision boundary; however, the same does not hold for generative tasks.

To address these gaps, we introduce the setting of Generative UDA, where an autoregressive model is trained to leverage knowledge from a labeled source domain to an unlabeled target domain, \textit{using only next word prediction as its objective}. We formalize the use of~\mlm for this setting in Section~\ref{sec:methodology}, and then attempt to explore and understand the behaviour of~\mlm for Generative UDA. 
We begin by performing an empirical analysis on 40 real-world domain pairs to explore the tradeoff between continued pre-training and invariance-based approaches, and find vanilla~\mlm to be competitive with and significantly more stable than a state of the art invariance-based approach (Section~\ref{sec:empirical-results}). We then stress test~\mlm, by applying it across varying model architectures and scales, tuning approaches and data regimes. We find that~\mlm is robust across these settings, unlike our invariance-based approach (Section~\ref{sec:results-more-settings}).

With recent language models being trained across vast corpora which may include domains similar to the target domain, the requirement for continued pre-training may be raised to question. In Section~\ref{sec:cpt-requirement}, we show that continued pre-training is indeed essential for strong downstream performance on the target domain, and this performance rapidly degrades with limited target domain exposure.
Finally, we attempt to shed light on how masking plays a role in improving classification accuracy on the unlabeled target domain in Section~\ref{sec:mask-word-analysis}.
We find that the model may implicitly learn the downstream task as it predicts masked words that are informative to the downstream task. 

Our work attempts to connect the body of UDA research with recent trends in language modeling, by providing a set of insights into the behaviour of the popular class of continued pre-training approaches, in the Generative UDA setting. We hope this enables an initial step towards a wider applicability of modern language models. 

\section{Continued Pre-Training for Generative UDA}
\label{sec:methodology}

\subsection{Preliminaries: The UDA Problem}
\label{sec:setup}

We consider a text classification task, where $\mathcal{X}$ is the input space of all text sentences and $\mathcal{Y} = \{1,...K\}$ is the label space. In the UDA problem, we have access to a source labeled dataset $\mathcal{D}_\text{src}=\{(x_i, y_i)\}_{i=1}^N$ consisting of samples from a joint distribution $P_{\text{src}}$, and a target unlabeled dataset $\mathcal{D}_\text{tgt}=\{x_j\}_{j=1}^M$ sampling from a target input distribution $P^{\mathcal{X}}_{\text{tgt}}$. We further denote $P^{\mathcal{X}}_{\text{src}}$ as the marginal distribution of $P_{\text{src}}$ on $\mathcal{X}$, where $P^{\mathcal{X}}_{\text{src}} \neq P^{\mathcal{X}}_{\text{tgt}}$. 
The goal of UDA is to learn a function $f: \mathcal{X} \to \mathcal{Y}$ that minimizes the error rate 
$\mathbb{E}_{x \sim P^\mathcal{X}_{\text{tgt}}} \; \mathbbm{1} [ f(x) \neq y]$.
\subsection{\mlm for UDA as a Sequence of Prompt Based Tasks}

We now formalize the extension of continued pre-training to the setting of generative UDA.
We use the traditional two-phase training pipeline from~\citet{gururangan2020don}\footnote{While we use the two-phase multi-task training pipeline (sequential) in our main experiments, in Appendix~\ref{sec:single-phase-mlm}, we show that an equivalent single-phase multi-task training pipeline (joint) results in similar performance.}.
The first phase uses templates to cast the source and target domain sequences into an autoregressive pre-training task.\footnote{We investigate mask language modeling for T5 models and switch to causal language modeling for decoder-only models with a few simple template changes. We compare both in Section~\ref{sec:results-more-settings}.}
The second phase applies supervised instruction-tuning of the model on source-labeled data.

\paragraph{Task 1: Autoregressive Continued Pre-training} 
We reuse the input sequences from the source-labeled dataset $\mathcal{D}_\text{src}$ as the source-unlabeled dataset, denoted as $\mathcal{D}_\text{src}^\mathcal{X}$. 
Next, similar to~\citet{raffel2020exploring, song2019mass}, for an unlabeled sequence  $x\in \mathcal{D}_\text{src}^\mathcal{X}$ and $\mathcal{D}_\text{src}$, we use a prompt template to convert the sequence $x$ to an input-output sequence pair, i.e., $\mathbb{M}(x) = (\Tilde{x}, \Tilde{y})$. 
For masked language modeling (MLM), an instruction is pre-pended to a randomly masked sequence $x$ to create $\Tilde{x}$. The output sequence $\Tilde{y}$ is a concatenation of masked words from $x$. 
For example, given $x=$\textit{``The movie was so cool! Two hours of fun.''}, we construct $\Tilde{x}=$\textit{``Fill in the blanks: "The \_ cool! Two hours \_''}, and $\Tilde{y}=$ \textit{``\texttt{<sep>} movie was so \texttt{<sep>} of fun. \texttt{<sep>}''}.
For causal language modeling (CLM), $\Tilde{x}=x$.

Given $(\Tilde{x}, \Tilde{y})$, we train an autoregressive LM parameterized by $\theta$ to minimize the negative log-likelihood loss averaged over output words and the total loss over a corpus $\mathcal{D}=\mathcal{D}_\text{src}^\mathcal{X} \cup \mathcal{D}_\text{tgt}$.
\begin{equation} \label{equn:cross-entropy}
    \ell(\Tilde{x}, \Tilde{y}; \theta) = 
    - \frac{1}{|\Tilde{y}|}\sum_t \log  P_\theta (\Tilde{y}_t | \Tilde{x}, \Tilde{y}_{1:t-1})
\end{equation}
\vspace{-0.2in}
\begin{align*}
    \mathcal{L}_\text{CPT}(\mathcal{D}; \theta) = 
    {\frac{1}{|\mathcal{D}|}} \sum_{x \in \mathcal{D}} \ell(\mathbb{M}(x); \theta)
\end{align*}

\paragraph{Task 2: Source Supervised Instruction-tuning}
In the second phase, we use labeled data from the source domain to train the model on the downstream classification task. 
Similar to the first phase, we use prompts\footnote{Prompt templates were selected from the Public Pool of Prompts \citep{bach2022promptsource}.} to generate input-output sequence pairs: $\mathbb{C}(x, y) = (\Tilde{x}, \Tilde{y}) \; \forall \; (x, y) \in \mathcal{D}_\text{src}$.
For example, for sentiment classification, if $x=$\textit{``I like this movie.''}, $y=1 \Rightarrow$  $\Tilde{x} =$ \textit{``[$x$] Is this sentence positive or negative?''}, $\Tilde{y} =$ \textit{``Positive''}.

Given the augmented sequence pair $(\Tilde{x}, \Tilde{y})$ and the model trained after the first phase, we compute the same negative log-likelihood loss $\ell(\Tilde{x}, \Tilde{y}; \theta)$ in Eq.~(\ref{equn:cross-entropy}). 
Finally, we define the total loss on the source-labeled dataset in the second phase as:
\begin{equation}  \label{equn:cls_loss}
    \mathcal{L}_\text{CLS}(\mathcal{D}_\text{src}; \theta) = {\frac{1}{|\mathcal{D}_\text{src}|}}\sum_{(x, y) \in \mathcal{D}_\text{src}} l(\mathbb{C}(x, y); \theta)
\end{equation}

After training, we follow the practice of~\citet{liu2022few} to convert a label string $\Tilde{y}$ to its corresponding label $y$ at test time for evaluation.

\section{Evaluating the Efficacy of Continued Pre-training for Generative UDA}
\label{sec:empirical-results}

\subsection{Experimental Setup}
\label{sec:experiments}

\paragraph{Datasets}
We use the MNLI and Amazon Review classification datasets, which are widely used UDA benchmarks~\citep{malik2023udapter, karouzos2021udalm, guo2020multi}.
The MNLI corpus \citep{williams2018broad} contains sentence pairs across five genres: Travel (T), Fiction (F), Government (G), Slate (S), and Telephone (Te). The task classifies every sentence pair as entailment, neutral, or contradiction. 
The Multi-Domain Sentiment Analysis Dataset~\citep{blitzer2007biographies} contains binary sentiment reviews for different types of Amazon products. We use reviews from the Apparel (A), Baby (B), Books (Bo), Cameras (C), and Movies (M) domains. 
We evaluate a total of 40 pairs of source and target domains, across the two datasets.
Appendix~\ref{sec:appendix-dataset-setup} contains more details about the datasets.

\paragraph{Models and Tuning Methods} 
Our main experiments use the T5v1.1 base model and (IA)$^3$~\citep{liu2022few} PEFT method. 
T5v1.1 is an improved version of the original T5 model \citep{raffel2020exploring}, and unlike the original T5 model, it is not trained on any supervised datasets.
We then extend our evaluation to different model architectures (T0, GPT-2), tuning methods (full fine-tuning, adapters) and data regimes (Section~\ref{sec:results-more-settings}).

\paragraph{Training} 
Each training phase is 30,000 steps long for MNLI and 15,000 steps for the Amazon dataset. We use Adam, a batch size of 8, and a learning rate of 0.003. We set the maximum sequence length to 256 tokens. 
We use length normalization during evaluation, as proposed by \citet{liu2022few}. 
For each experiment, we report the mean and standard deviation across 3 runs.
More details can be found in Appendix~\ref{sec:appendix-implementation-details}. 

\paragraph{Baselines}
Since our goal is to study the behaviour of~\mlm for generative UDA, we compare it with a simple supervised baseline, and a state-of-the-art invariance-based approach.
\begin{itemize}[leftmargin=12pt,topsep=1pt] \itemsep-0.2em
\item \textbf{\supervised} (All labeled): We fine-tune the model using labeled data from both the source and target domains. This serves as an upper bound on target domain performance. 
\item \textbf{\mmd}: \citet{malik2023udapter} propose an invariance-based method that measures the multi-kernel maximum mean discrepancy (MK-MMD) \citep{gretton2012kernel, bousmalis2016domain} between source and target embeddings from each transformer layer and sums them to obtain an aggregate loss $\mathcal{L}_\text{div}$. The final loss is the weighted sum of $\mathcal{L}_\text{div}$ and the classification loss, i.e., $\mathcal{L} = \lambda \; \mathcal{L}_\text{cls} + (1 - \lambda) \; \mathcal{L}_\text{div}$, where $\lambda$ gradually changes from 0 to 1 during training. 
Here, we use the embeddings from a model as it is being instruction tuned on the downstream classification task.
Their method achieves state-of-the-art performance, and outperforms popular UDA approaches like DANN~\citep{ganin2016domain} and DSN~\citep{bousmalis2016domain}; thus we only compare~\mlm with this approach. 
\end{itemize}

\subsection{Continued Pre-training is Competitive with Domain-Invariance Methods}
\label{sec:main-results}

\paragraph{Performance} 
We compare~\mlm with other baselines over 40 domain pairs of the MNLI and Amazon Review datasets, and report the average accuracies over all pairs in Table~\ref{tab:results-combined-short}.
We see that \mlm~is competitive to \mmd. (Appendix~\ref{sec:appendix-main-results-amazon-mnli} contains detailed results over 40 pairs and significance tests to check for competitiveness.)
Interestingly, a visualization of sentence embeddings in Figure~\ref{fig:mlm-understanding} (Appendix~\ref{sec:appendix-main-results-amazon-mnli}) suggests that representations learned through~\mlm are not domain invariant. In addition to the MMD based method of~\citet{malik2023udapter}, we also compare~\mlm on one domain pair with other methods that promote domain invariance (DANN~\citep{ganin2016domain}, CORAL~\citep{sun2017correlation}) and weight interpolation~\citep{ilharco2022editing} in Appendix~\ref{sec:appendix-main-results-amazon-mnli}, further confirming the competitiveness of CPT.

\begin{table}[!ht]
    \centering
    \vspace{-1mm}
    \resizebox{7cm}{!}{
    \begin{tabular}{llll}
    \toprule
    \textbf{Dataset} & \textbf{\supervised} & \textbf{\mmd} & \textbf{\mlm} \\
    \midrule
    Amazon & 92.66 (0.45) & 89.02 (2.17) & \textbf{89.34} (0.48) \\
    MNLI & 78.14 (0.25) & 70.19 (1.71) & \textbf{74.12} (0.68) \\
    \bottomrule
    \end{tabular}
    }
\caption{Avg. target-domain classification accuracy and standard deviation over 3 runs.}
\label{tab:results-combined-short}
\end{table}

\vspace{-0.1in}
\paragraph{Stability} 
\mlm performs more stably than \mmd, with the invariance-based method often reporting a variance of over 20\% across runs (Table~\ref{tab:results-combined-no-src-only} in Appendix~\ref{sec:appendix-main-results-amazon-mnli}).
For example, for the MNLI pair Fiction (F) $\to$ Government (G), minimizing \mmd~yields a variance of 23.4\% across runs.
This observation is consistent with existing findings~\citep{ramesh2021domain, han2019unsupervised} that minimizing divergence measures like MMD, when combined with auxiliary task-specific loss functions, result in training instabilities and vanishing gradients. 
We discuss this in more detail in Appendix~\ref{sec:mmd-variations}.

\section{How General are the Benefits of~\mlm?}
\label{sec:results-more-settings}

Instruction tuning for large models is often performed through parameter-efficient fine-tuning (PEFT) on limited data. This tuning also applies to models of different scales and architectures (decoder-only and encoder-decoder). In this section, we evaluate the utility of~\mlm across these factors, using the A$\to$M domain pair from the Amazon Reviews dataset.

\paragraph{\mlm Helps Decoder-only Models.} We extend our analysis from MLM with encoder-decoder language models, to causal language modeling (CLM) with decoder-only language models, using GPT-2 (medium)~\citep{radford2019language}.
We perform CLM in the first training phase by simply training the model for next-word prediction given the original sequence. 
Table~\ref{tab:decoder-models}
shows that \mlm~provides strong improvements on the target domain
in comparison to the invariance-based baseline.

\begin{table}[!ht]
    \centering
    \resizebox{3.5cm}{!}{
    \begin{tabular}{ll}
    \toprule
    \textbf{Method} & \textbf{Accuracy} \\
    \midrule
    \supervised & 79.8 (0.3) \\
    \mmd & 66.0 (1.1) \\
    \mlm & \textbf{75.8} (0.4)\\
    \bottomrule
    \end{tabular}
    }
\caption{Performance of \mlm~with causal language modeling for the decoder-only GPT-2 model. \mlm significantly outperforms the invariance-based method.
}
\label{tab:decoder-models}
\end{table}

\vspace{-0.15in}
\paragraph{\mlm Outweighs Invariance-based Methods for Instruction-tuned Models.}
We evaluate the performance of \mlm~over T5v1.1 XL (3B parameters) and the instruction-tuned T0 (3B parameters)~\citep{sanhmultitask}. 
Figure~\ref{fig:ins-tuned-models} (Table~\ref{tab:ins-tuned-models} of Appendix~\ref{sec:appendix-other-models})
shows a wider gap between \mmd~and \mlm~with higher model capacity, and this gap is further increased with instruction tuning.
We hypothesize that this gap is due to the vast difference between the objectives of domain invariance and instruction tuning.

\begin{figure}[htbp]
  \centering
    \centering
    \begin{center}
    \centerline{\includegraphics[width=0.8\columnwidth]{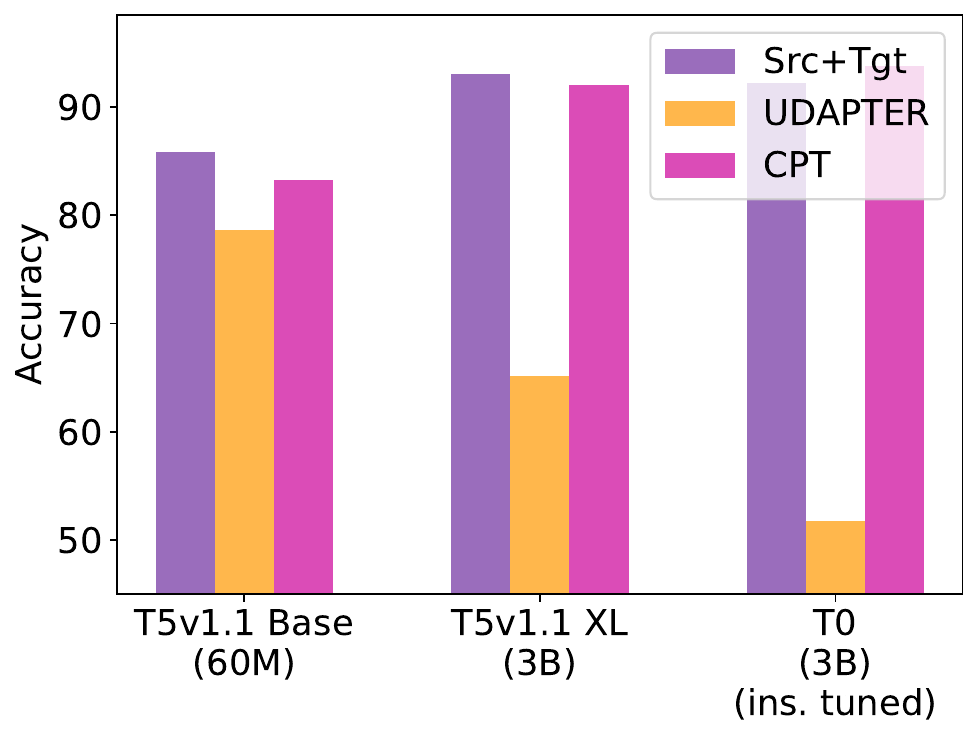}}
    \caption{The performance gap between \mlm~and \mmd~increases with larger models, from T5v1.1 Base (60M parameters) to T5v1.1 XL (3B parameters), and further increases with instruction tuning (T0 3B).}
    \label{fig:ins-tuned-models}
    \end{center}
\end{figure}

\vspace{-0.25in}
\paragraph{\mlm Benefits are Consistent across Tuning Approaches.}
PEFT approaches have been shown to introduce resilience to domain shift~\citep{fu2023effectiveness}. To isolate this effect from the~\mlm framework, we use T5v1.1 to evaluate~\mlm in a full fine-tuning setup. Additionally, we compare \mlm with two PEFT approaches\footnote{We choose Adapters because \citet{hetowards} present a unified view of PEFT approaches which shows that the operations applied by Adapters are very similar to those of 
Prefix Tuning \citep{li2021prefix} and LoRA \citep{hulora}. We choose (IA)$^3$ since it is a state-of-the-art PEFT approach that uses a fraction of the learnable parameters of Adapters (More in Appendix~\ref{sec:appendix-peft-frameworks}).} :
Adapters~\citep{houlsby2019parameter}
and (IA)$^3$~\citep{liu2022few}.
We see in Figure~\ref{fig:varying-peft} (Table~\ref{tab:varying-peft} in Appendix~\ref{sec:appendix-peft-frameworks}) that~\mlm continues to perform stronger than the domain invariance-based~\mmd method.

\begin{figure}[ht]
    \centering
    \includegraphics[width=0.8\columnwidth]{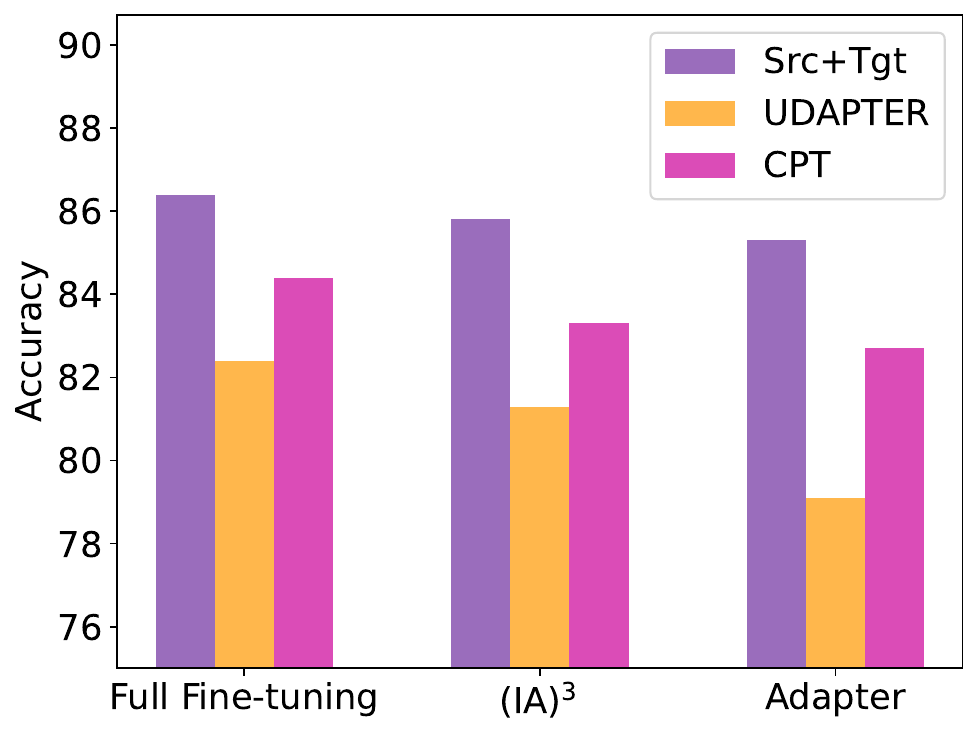}
    \caption{Performance of \mlm~across different tuning approaches with the T5v1.1 base model. \mlm~remains more powerful than \mmd~across all tuning approaches.}
    \label{fig:varying-peft}
\end{figure}

\vspace{-0.15in}
\paragraph{\mlm Outperforms Invariance-based Methods in Low-data Regimes.}
In this low-data experiment, we assume access to $k$ labeled source-domain examples. 
For~\mlm, we assume access to the full unlabeled dataset in both domains
for the first training phase, and $k$-shot access to labeled source-domain examples for the second phase of supervised training. 
For a fair comparison, we also introduce a two-phase version of the~\mmd pipeline---the first phase minimizes MMD between unlabeled source and target domain embeddings (full data access), while the second phase optimizes supervised training on the source domain ($k$-shot). 
Figure~\ref{fig:256-shot-ins-tuned-models} (Table~\ref{tab:k-shot} in Appendix~\ref{sec:appendix-few-shot-learning}) showcases \mlm~clearly outperforming both variants of~\mmd, across three different models for $k=256$.
Furthermore, Figure~\ref{fig:k-shot} (Table~\ref{tab:performance-across-shots} in Appendix~\ref{sec:appendix-few-shot-learning}) shows~\mlm providing consistent improvements in as low as 32-shots, unlike the unstable invariance-based approach.

\begin{figure}[htbp]
  \centering
    \begin{center}
    \centerline{\includegraphics[width=0.8\columnwidth]{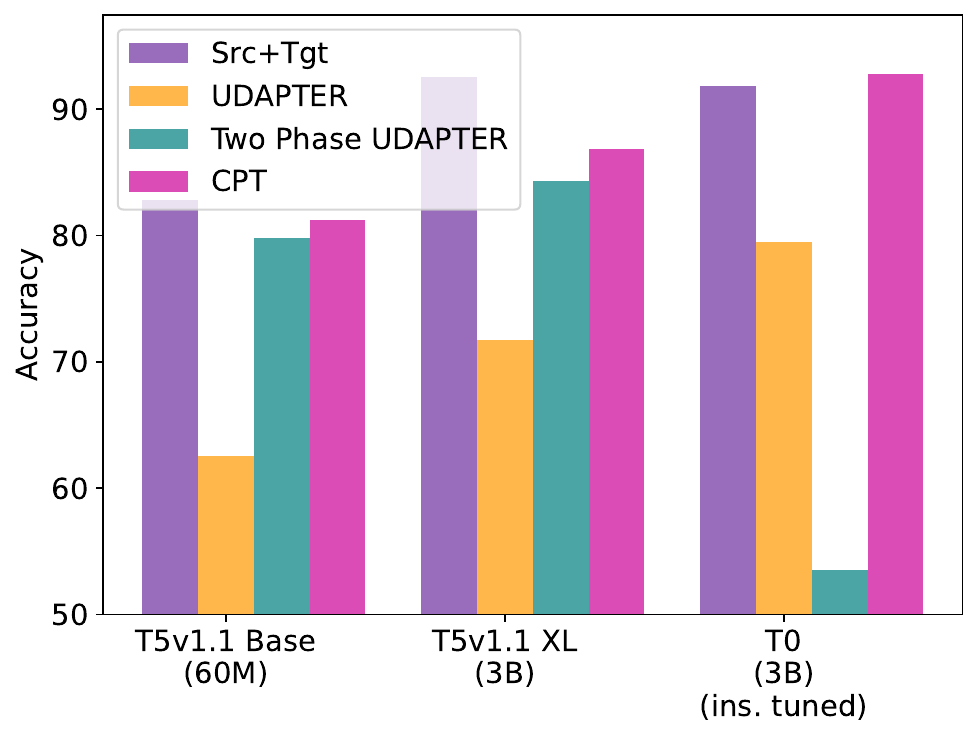}}
    \vskip -0.1in
    \caption{Performance of \mlm across different models, in a 256-shot learning setup. Unlike both variants of~\mmd,~\mlm is stable and provides consistent benefits across models.}
    \label{fig:256-shot-ins-tuned-models}
    \end{center}
  \vskip -0.3in
\end{figure}

\section{To What Extent is Target-Domain Exposure Beneficial?}
\label{sec:cpt-requirement}

Given the vast distributions language models are pre-trained on, a natural assumption might be that the model has already been exposed to a domain similar to the target domain during pre-training. This would mean that the model could simply extrapolate the learned downstream task from the source to the target domain, questioning the need for CPT.

In this section, we establish that exposure to the target domain \textit{is} helpful, even when similar domains may have been encountered during pre-training. 
Table~\ref{tab:phase1-ablation} evaluates~\mlm on the A$\rightarrow$M domain pair with the T5v1.1 model. 
We note that the performance on the target domain is strongly impacted by the presence of target-domain data during the first phase of training. 

\begin{table}[!ht]
    \centering
    \resizebox{0.7\columnwidth}{!}{
    \begin{tabular}{lll}
    \toprule
    \multirow{2}{*}{\textbf{Phase 1 Data}} & \multicolumn{2}{c}{\textbf{Accuracy}} \\     
    ~ & \multicolumn{1}{c}{Source} & \multicolumn{1}{c}{Target} \\ 
    \midrule
        Source Only & 93.3 (0.1) & 76.5 (0.2)\\
        Target Only & 92.9 (0.4) & 	82.3 (0.7) \\
        Source + Target & \textbf{93.5} (0.4) & \textbf{83.3} (0.5) \\
    \bottomrule
    \end{tabular}
    }
    \captionof{table}{Comparison of \mlm~with varying data exposure during the first phase of training. Performance on the target domain strongly benefits more from exposure to target domain, and is boosted further with exposure to the source domain.
    }
    \label{tab:phase1-ablation}
\end{table}

\begin{figure}[htbp]
  \centering
    \begin{center}
    \centerline{\includegraphics[width=0.8\columnwidth]{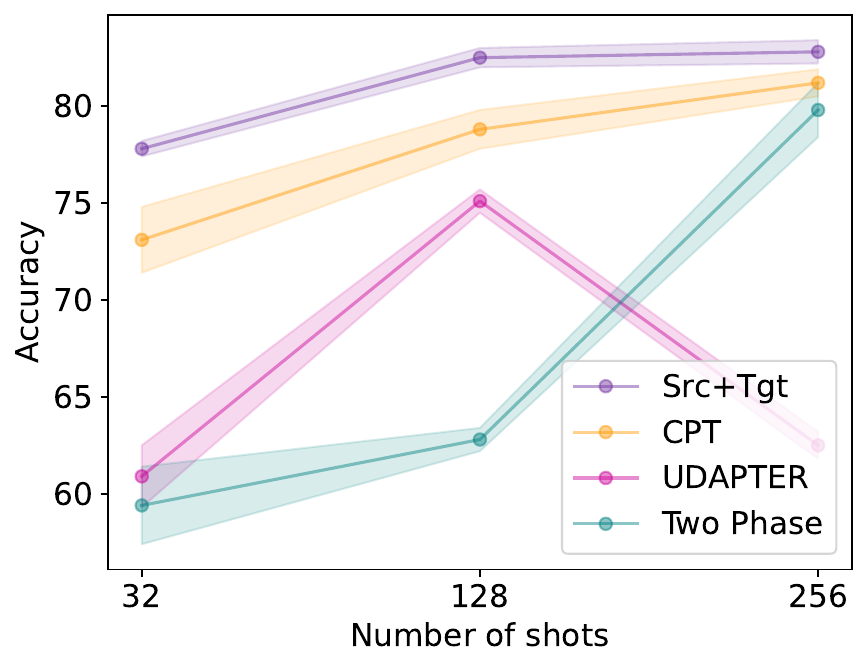}}
    \caption{Few-shot performance of~\mlm, for varying $k$. Unlike both variants of~\mmd,~\mlm is stable and provides consistent benefits across number of shots.}
    \label{fig:k-shot}
    \end{center}
  \vskip -0.25in
\end{figure}

For a more fine-grained analysis, we investigate the impact of degree of exposure to the target domain, by varying the masking rates during the first phase of training.
While masking 15\% of a sequence is considered standard for random masking, previous work has shown that BERT-sized models \citep{delvin2019bert} can learn from as high as 80\% masking rates during pre-training followed by adaptation to a labeled task~\citep{wettig2022should}. The source-domain performance shown in Figure~\ref{fig:mask-rate-impact} (Table~\ref{tab:masking-rate} in Appendix) matches this trend. However, high masking rates effectively reduce the exposure of the model to target data, strongly deteriorating the performance on the target domain\footnote{With masking rates under the optimal value of 15\%, the semantic and background features learned through model prediction of masked words is limited, hurting performance on the target domain.}. We hypothesize that since the model never sees any labeled data of the target domain, it heavily depends on the signal it gets from the unlabeled data through masking. 

\begin{figure}[htbp]
  \centering
    \begin{center}
    \centerline{\includegraphics[width=0.8\columnwidth]{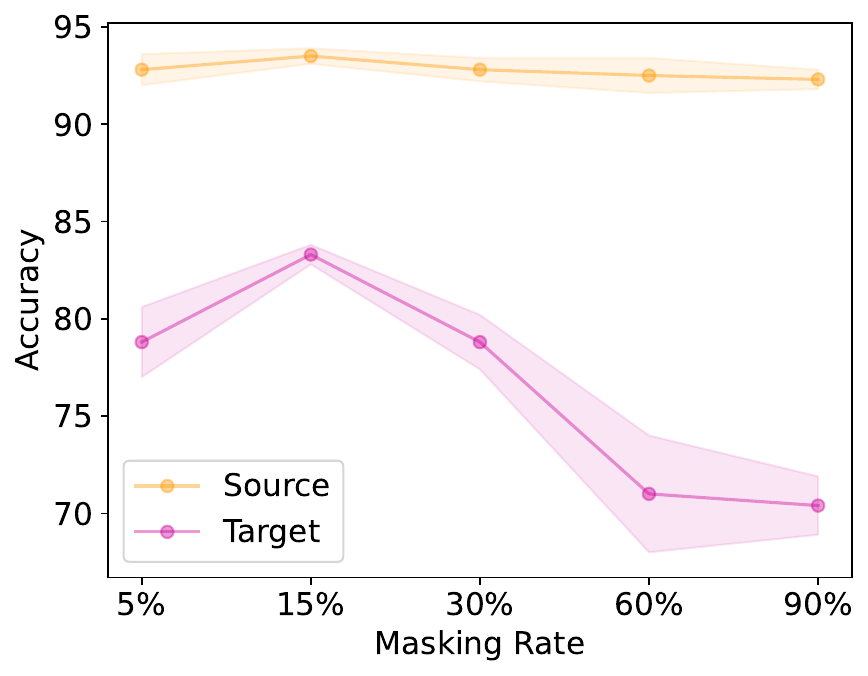}}
    \caption{Impact of Masking Rate on \mlm.
    With high masking rates, the performance on the source domain is largely maintained, but the performance on the target domain rapidly deteriorates.}
    \label{fig:mask-rate-impact}
    \end{center}
\end{figure}

\vspace{-0.2in}
\section{Why does Word Prediction Aid Classification for Generative UDA?}
\label{sec:mask-word-analysis}

In this section, we examine why predicting masked words of the source and target domains through~\mlm boosts sentence classification on the unlabeled target domain for generative UDA. 
We hypothesize that by having to predict masked words that are informative to the downstream task during pre-training, the model implicitly learns information about the downstream task. 
For example, given the masked sentence,
\textit{``I really \_ the movie, it was a fascinating watch.''}, 
the masked word is indicative of the downstream task, in this case sentiment analysis. The model can only predict this masked word (which would be a positive word like ``\textit{loved}'' or ``\textit{enjoyed}'') by using other words informative to the task (``\textit{fascinating}''). Through this process, the model is essentially learning features which are useful to the downstream task, despite having no direct supervision.

To test this hypothesis, we quantize the ``informativeness'' of each word to a classification task: an informative word is highly correlated with any of the labels in the downstream task.\footnote{These informative words are similar to pivot features~\citep[\textit{inter alia}]{blitzer2006domain, ziser2018pivot, ben2020perl}, with the exception that they are chosen based on information from the source domain only.}
Specifically, we follow \citet{gururangan2018annotation} and use pointwise mutual information (PMI) \citep{fano1961transmission} of the word with respect to the class label:
\begin{equation*}
    \text{PMI}(\text{word}, \text{class}) = \log \frac{p(\text{word}, \text{class})}{p(\text{word}) p(\text{class})},
\end{equation*}
where we count the frequency of a word-class pair on $\mathcal{D}_\text{src}$ to estimate $p(\text{word}, \text{class})$, and similarly count a word and a class individually on $\mathcal{D}_\text{src}$ to estimate $p(\text{word})$ and $p(\text{class})$.

\begin{figure}[htbp]
  \centering
    \centering
    \begin{center}
    \centerline{\includegraphics[width=0.5\columnwidth]{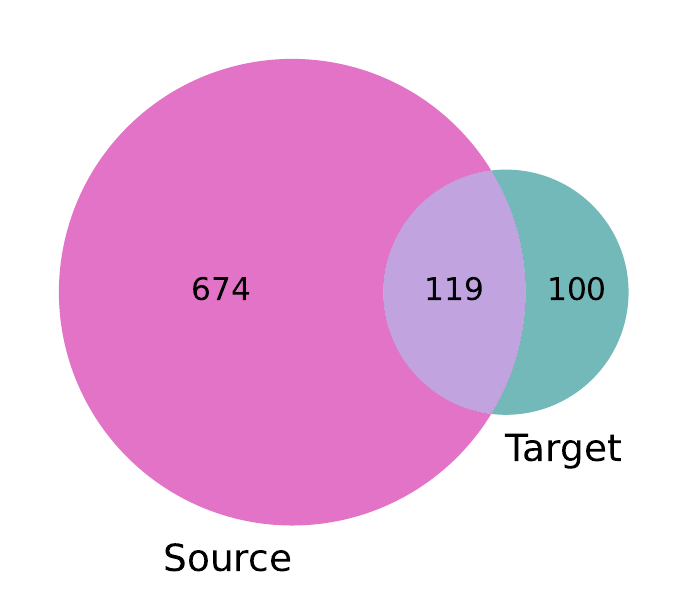}}
    \caption{Vocabulary overlap between label-informative words of the source and target domains. The numbers in the Venn diagram indicate the number of words in both sets.}
    \label{fig:info-word-overlap}
    \end{center}
\vskip -0.15in
\end{figure}

We use two sets of words from a dataset: those with the top $k\%$ (informative) and bottom $k\%$ (uninformative) PMI with any inference label ($k=15$). 
We also filter out low-frequency words from the selection.\footnote{Any word that occurs less than 10 times in the entire training corpus is considered to be low frequency.}
We compute these sets for the source and target domains individually, assuming access to target labels.
We use the T5v1.1 model on the A$\to$M pair for our analysis.

\begin{figure*}[htbp]
    \centering
    \begin{center}
    \centerline{\includegraphics[width=0.8\columnwidth]{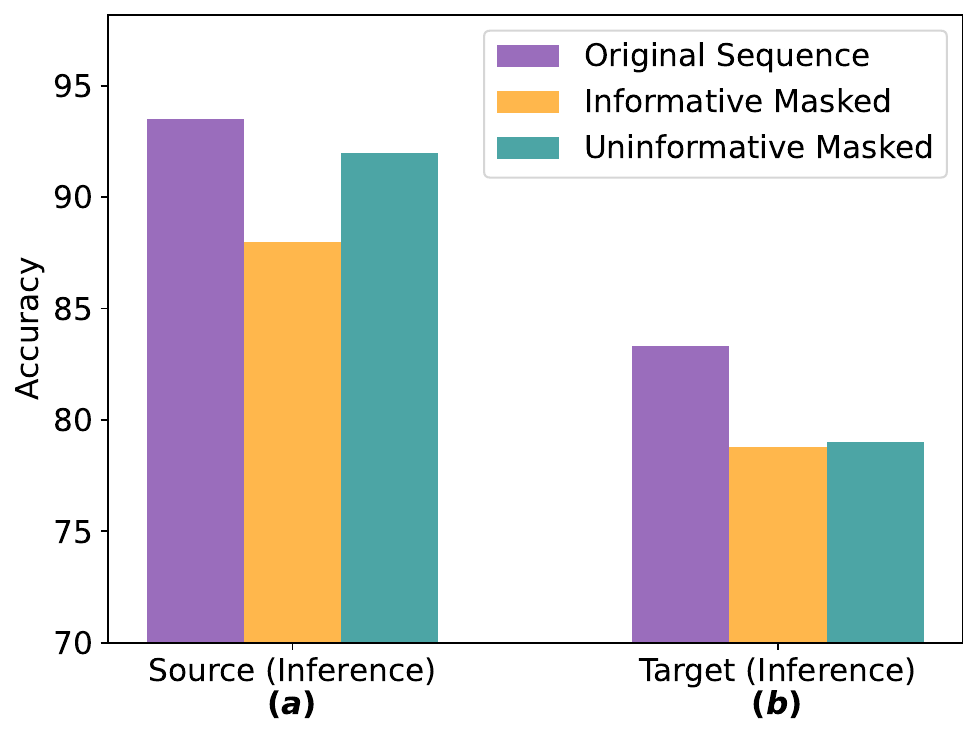}
    \includegraphics[width=0.8\columnwidth]{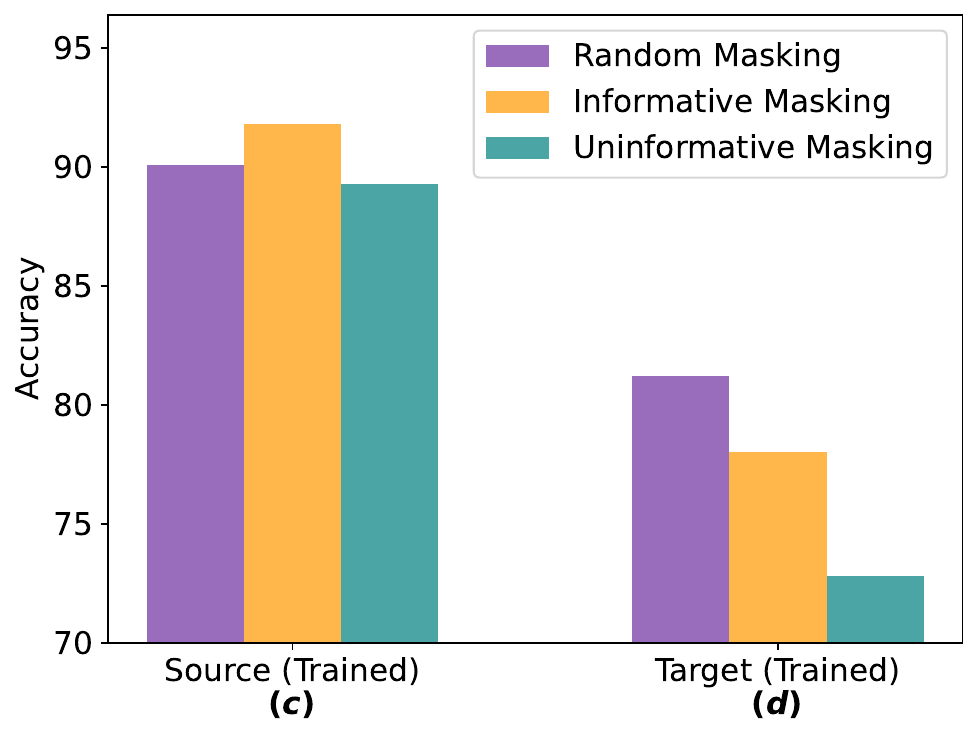}}
    \caption{The impact of selective masking on classification performance of a~\mlm trained model. 
    \textbf{Left}: Masking label informative words during inference degrades classification performance, compared to the original unmasked sequence. However, the removal of uninformative words does not impact the model on the source domain.
    \textbf{Right}: The impact of continued pre-training can be partially disentangled from that of supervised training by making the supervised training phase few-shot and training the model to mask informative or uninformative words during the continued pre-training phase. Informative masking is most beneficial for the source domain, indicating that the model learns task relevant features during masking. On the target domain, informative masking still captures some knowledge about the downstream task, however, the supervised training phase adds a bias towards source label informative words. Thus, random masking is most powerful.
    }
    \label{fig:info-vs-uninfo-masking}
    \end{center}
\end{figure*}

\paragraph{How Masking helps Learn Classification}
We first confirm that label-informative words indeed impact the classification performance of the~\mlm model. 
We do this by masking informative words from a sentence at inference. 
Figure~\ref{fig:info-vs-uninfo-masking} (a) shows us that the performance of the model on the source domain is not impacted by masking uninformative words, but drops on masking informative words. 
However, how do we know how much of this bias towards label informative words was learned during the continued pre-training phase, rather than during supervised fine-tuning? To attempt to disentangle the impact of the training phases, we train the model through selective masking (informative or uninformative) in the first phase of training, and minimize the impact of the second phase by making it a few-shot task.
Figure~\ref{fig:info-vs-uninfo-masking} (c) shows us that the model performs best on classification when trained to predict label informative words during masking, 
indicating that the model does indeed learn features relevant to the downstream task during the first phase of training.

\paragraph{The Interplay between CPT and Classification for Generative UDA} 
We now extend this analysis to the target domain to understand how~\mlm plays a role in learning features from the unlabeled domain.
Figure~\ref{fig:info-vs-uninfo-masking} (d) shows us that informative masking outperforms uninformative masking by a significant gap, once again signaling that the masking process helps the model implicitly learn the downstream task. 
However, unlike with the source domain, random masking results in the strongest performance. This is due to the domain mismatch: the informative words for the source and target domains are not identical (Figure~\ref{fig:info-word-overlap}), and the supervised training on the source domain adds a bias towards source-informative words. 
The mixture of these two sets of words are best predicted through random masking, explaining its strong performance.

This phenomenon also draws the observation that random masking is preferred to selective masking for generative UDA, contrary to single domain settings where informative masking is more useful~\citep{levinepmi, gu-etal-2020-train}.

\section{Discussion}
\label{sec:discussion}

\paragraph{The Computational Trade-off of~\mlm} Our results in Section~\ref{sec:main-results} show that continued pre-training and methods promoting domain invariance are competitive with each other. Continued pre-training suffers from the computational drawback of requiring an additional phase of training. Conversely, invariance-based methods are difficult to optimize, possibly requiring more runs to achieve a stable optimum, and having a higher amortized computational cost. Inspired by~\citet{karouzos2021udalm}, we introduce a simple single phase variant of continued pre-training which is \textit{equivalent} in performance to its two phase variant, nullifying the additional computational overhead of the approach (more details in Appendix~\ref{sec:single-phase-mlm}).

\paragraph{UDA in the Age of LLMs}
Recent breakthroughs in scale have showcased that large language models (LLMs) are highly powerful, and can perform various downstream tasks with limited or no training. This may raise question on the relevance of the UDA problem as a whole --- does a model even require expensive adaptation to a domain it may have already been exposed to during pre-training?
In addition to our analysis in Section~\ref{sec:cpt-requirement}, 
we argue that the requirement to adapt small models to unseen domains still holds in specific cases. 
Small supervised models have been shown to be comparable with, or even outperform, zero-shot general-purpose LLMs on various downstream tasks~\citep{huang2023reevaluation, zhu2023multilingual, tang2024minicheck}, serving as lightweight and customizeable (through fine-tuning) alternatives. Safety critical domains like healthcare and finance would benefit more from these models than a generalist LLM. 
Our study does not address how to better adapt to domains, rather we investigate ways a model may adapt to data unseen during pre-training. This is a question that holds for current LLMs, and will continue to hold as long as models are unable to access infinite data during pre-training.

\section{Related Work}
\label{sec:related-work}

\paragraph{UDA through Promoting Domain Invariance}
A major class of approaches in Model-centric UDA methods~\citep{ramponi2020neural} aims to minimize $\mathcal{H}\Delta\mathcal{H}$ divergence~\citep{ben2010theory} between the source and target domain features, through adversarial training \citep[\textit{inter alia}]{tzeng2014deep, ganin2016domain, tzeng2017adversarial, 
guo-etal-2022-improving} or through minimizing measures of domain similarity~\citep{bousmalis2016domain, ge2023unsupervised}.
\citet{malik2023udapter} have shown the minimization of MMD to outperform other invariance-based methods.
However, past work has shown that domain-invariance is a 
weak constraint for adaptation~\citep{zhao2019learning, karouzos2021udalm}, 
could introduce domain-specific hyperparameters~\citep{trung2022unsupervised},
and is also prone to instability issues~\citep{han2019unsupervised, sun2019unsupervised, wilson2020survey, ramesh2021domain}.

\paragraph{UDA through Continued Pre-Training} 
The limitations of invariance-based model-centric methods have encouraged the emergence of alternate approaches, based on self-supervised learning through contrastive learning~\citep{kumarfine, shen2022connect, long2022domain}, pseudo-labeling~\citep[\textit{inter alia}]{zhou2005tri, ruder2017learning} or language model pre-training.
Despite not being directly useful to certain downstream tasks \citep{uppaal2023fine},~\mlm has been used for adaptation to labeled tasks, in both full fine-tuning~\citep[\textit{inter alia}]{gururangan2020don, lee2020biobert, gao2021making} and PEFT setups~\citep{kim2021revisiting, hung2023tada}. 
A smaller body of work has explored the utility of~\mlm in a UDA setup~\citep{han2019unsupervised, zhangunsupervised, karouzos2021udalm, parovic2023cross}, identifying the 
class of methods to be more stable than invariance-based methods.

\paragraph{Generative UDA}
The emergence of large language models~\citep[\textit{inter alia}]{brown2020language, scao2022bloom, touvron2023llama} introduced the concept of instruction tuning with templates~\citep{zhang2023instruction, sanhmultitask, ouyang2022training, wang2022super, weifinetuned, gao2021making, liu2023pre}, enabling multi-task training without any task specific architectural changes.
However, the framework of casting discriminative classification tasks into generative next word prediction tasks has not yet been extended to UDA.
The closest work to this setting~\citep{ben2021pada} uses a generative model to create domain identifier prompts and feed them back into the model, however the final task label prediction is still discriminative. 
In our work, we focus on gaining insights to extend the powerful class of~\mlm methods to purely generative UDA, where prediction on the downstream task is treated as a next word prediction task. 
Through this, we also present novel findings on the impact of~\mlm to prompt-based classifiers in the UDA framework, countering previous findings from other studies in a single-domain setting~\citep{gu-etal-2020-train, levinepmi, wettig2022should}.

\section{Conclusion}
\label{sec:conclusion}

\vspace{-0.1cm}
We introduce the setting of Generative UDA, and perform an investigation on the utility of continued pre-training in this setting. 
We compare the approach with the popular class of domain-invariance based methods for UDA, showing that~\mlm is both competitive with, and more stable than invariance-based approaches.
Our experiments show that the benefits of~\mlm extend to different architectures, tuning methods and data regimes.
We motivate the need for target domain exposure through~\mlm by showing that performance on the target domain gradually degrades with increasing masking rate.
Finally, we shed light on the interplay between masking and classification performance, and how this aids UDA. 
Our analysis shows that in predicting masked words that are informative to the downstream task, the model implicitly learns about the downstream task, furthering the benefits of directly learning the task.
Our work connects the body of UDA research with that of instruction tuning, enabling an initial step towards a wider applicability of modern language models.

\section*{Limitations}
\label{sec:limitations}

Our work presents an investigation into continued pre-training for UDA in a \textit{generative} setting. Since generative UDA is an almost completely unexplored area, we establish a proof of concept by using sentence classification tasks for our analysis. We leave the extending our analysis to more complex tasks to future work.

In our study, we consider a class of PEFT methods that involve inserting learnable parameters between the layers of the model. Other classes of PEFT methods were not considered.
However, we use Adapters and \citet{hetowards} have shown connections between the method with Prefix Tuning~ \citep{li2021prefix} and LoRA~\citep{hulora}.

Due to the high variance across runs in PEFT-based learning, we note that the performance can vary significantly across random seeds. We attempt to make our findings reproducible by averaging every experiment over 3 seeds.
Taking environmental costs into consideration, we reduce our computational budget by running a majority of our experiments with a smaller-sized model. Learning with larger models is discussed in Section~\ref{sec:results-more-settings}.

\section*{Ethics Statement}
\label{sec:ethical-considerations}

Our project aims to extend the problem of unsupervised domain adaptation to the generative setting, matching current needs with large language models. This is an effort towards improving the reliability and safety of language models, which can be fragile under distribution shift~\cite{ribeiro2020beyond} and incur great costs  over incorrect predictions \cite{ulmer2020trust, zhang2021out}. 

Our study does not involve any human subjects or violation of legal
compliance. We do not anticipate any potentially harmful consequences to our work. As detailed in Appendix~\ref{sec:appendix-dataset-setup}, all of our experiments are  conducted using publicly available datasets. Our code shall be released for reproducibility. Through our study and releasing our code, we hope to raise stronger research and societal awareness toward the problem of unsupervised domain adaptation in natural language processing.

\bibliography{anthology,custom}
\bibliographystyle{acl_natbib}

\clearpage
\appendix
\appendix

\section{Preparation of Evaluation Benchmarks}
\label{sec:appendix-dataset-setup}

We use two classification datasets, with 5 domains each. This results in a total of 40 pairs of source and target domains. For brevity, we include results of 24 domain pairs in the main paper, and the remaining 16 in Appendix~\ref{sec:appendix-main-results-amazon-mnli}. For both datasets, we use the train, validation and test splits from \cite{malik2023udapter}. More statistics about each dataset is available in Table~\ref{tab:dataset-details}. The listed datasets are intended for research purposes only. We do not make any commercial use of them. 

\paragraph{MNLI} The Multigenre Natural Language Inference (MNLI) corpus \cite{williams2018broad} contains sentence pairs across multiple genres: Travel (T), Fiction (F), Government (G), Slate (S) and Telephone (Te). The NLI task involves classifying every premise-hypothesis sentence pair as Entailment, Neutral or Contradiction. 

\paragraph{Amazon} The Multi Domain Sentiment Analysis Dataset \cite{blitzer2007biographies} contains Amazon product reviews for different type of products. We use reviews from the Apparel (A), Baby (B), Books (Bo), Cameras (C) and Movies (M) domains.  Each review is labelled as positive or negative. 

\begin{table}[ht]
\centering
\resizebox{7cm}{!}{
    \begin{tabular}{lllllll}
    \toprule
        \multirow{2}{*}{\textbf{Dataset}} & \multirow{2}{*}{\textbf{Language}} & \multirow{2}{*}{\textbf{License}} & \multicolumn{3}{c}{\textbf{Statistics per Domain}} \\ 
        ~ & ~ & ~ & Train & Val & Test \\ \midrule
        
    \texttt{MNLI} & English & cc-by-4.0 & 69600* & 7735** & 1945\\ 
        \texttt{Amazon} & English & cc-by-4.0 & 1440 & 160 & 400\\ 
        
        \bottomrule
    \end{tabular}
    }
\caption{Artifacts used in our study. The dataset statistics report the values used in our study.\\
* All domains contain approximately 69,600 examples. The exception is the Telephone domain, with 75,013 examples.\\
** All domains contain 7735 validation examples, except for Slate and Telephone, which contain 7731 and 8336 examples respectively.}
\label{tab:dataset-details}
\end{table}

\section{Details on Implementation}
\label{sec:appendix-implementation-details}

\paragraph{Models and Implementation}
We use T5v1.1, T0 and GPT-2 and LLaMA-2 from the HuggingFace library\footnote{\url{https://github.com/huggingface/transformers}}, and use PyTorch\footnote{\url{https://pytorch.org/}}
to train our models.

\paragraph{Training} We use the default hyperparameters from \citet{liu2022few}, except for batch size and training duration. We perform a grid search for these values. 
We train each training phase for 30,000 steps on MNLI and 15,000 steps on the Amazon dataset, with a batch size of 8. 
For the T5v1.1 XL and T0 models (3B parameters each), we use a batch size of 1.
We train with Adam and use a learning rate of 0.003. We set the maximum sequence length to 256 tokens. We use length normalization during evaluation, as proposed by \citet{liu2022few}. 
For each experiment, we report the mean and standard deviation across 3 runs.

We choose the number of training steps based on early stopping on the validation set for one domain, and use that number of steps for all domains within that dataset. We report the test set performance after a varying number of training steps in the table below. For example, for the Apparel$\rightarrow$Movies domain pair of the Amazon Reviews dataset, the performance saturates at 15,000 steps, as shown in Table~\ref{tab:training-duration}.

\begin{table}[t]
\begin{center}
\resizebox{7.5cm}{!}{
\begin{tabular}{lll}
\toprule
\multicolumn{1}{c}{\bf Training Steps} 
&\multicolumn{1}{c}{\bf Source Accuracy} 
&\multicolumn{1}{c}{\bf Target Accuracy}\\
\midrule
5,000    & 93.2 (0.4) & 81.9 (0.4) \\
10,000   & 93.4 (0.5) & 81.6 (0.6) \\
15,000   & 93.5 (0.4) & 83.3 (0.5) \\
\bottomrule
\end{tabular}
}
\end{center}
\caption{We use early stopping on one domain pair to determine the number of training steps, which we then use for all domain pairs of that dataset. For example, the Apparel$\rightarrow$Movies domain pair of the Amazon Reviews dataset shown in the table saturates at 15,000 steps.}
\label{tab:training-duration}
\end{table}

\paragraph{Computations} 
Using the (IA)$^3$ PEFT framework, training the T5v1.1 Base model (60 million parameters) for 15,000 steps takes approximately two hours on a single NVIDIA RTX A6000 GPU. 
The T5v1.1 XL model and T0 model (3 billion parameters) take approximately 8 hours for 15,000 steps of training.
For reproducibility, each experiment is repeated thrice, with changing random seeds.  
In total, we run 540 experiments with the Base model and 72 experiments with the larger models. This results in a total compute time of approximately 2400 GPU hours.





\section{Detailed results with the Amazon and MNLI Datasets}
\label{sec:appendix-main-results-amazon-mnli}

Table~\ref{tab:results-combined-no-src-only} shows the performance of \mlm on the Amazon and MNLI datasets. 

On the Amazon dataset, \mlm~is competitive with the state of the art \mmd~method from \citet{malik2023udapter} on average.
We confirm this by checking for a significant difference in the performance of \mlm and \mmd on the 20 dataset pairs. 
The Mann-Whitney U test and Student's t-test both resulted in non-significant p-values of 0.5516 and 0.8316, confirming the hypothesis that there is no significant difference between \mlm and \mmd on the Amazon dataset.

However, on the MNLI dataset, where all domains have larger gaps, both significant tests showed a significant difference between \mlm and \mmd, with \mlm being more powerful. This is exemplified through cases like Travel (T) $\to$ Government (G), where \mlm~yields an accuracy of 83.6\% on the target domain, equalling the upper bound of the \supervised~baseline. 

\begin{table*}[!ht]
    \centering
    \resizebox{0.8\textwidth}{!}{
    \begin{tabular}{llll|llll}
    \toprule
        \multicolumn{4}{c}{\textbf{Amazon}} & \multicolumn{4}{c}{\textbf{MNLI}} \\
        ~ & \textbf{\supervised} & \textbf{\mmd} & \textbf{\mlm} & ~ &\textbf{\supervised} & \textbf{\mmd} & \textbf{\mlm} \\ 
        \midrule
        A $\rightarrow$ B & 94.7 (0.2) & 93.8 (0.3) & \textbf{93.9} (0.3) & T $\rightarrow$ F & 77.2 (0.4) & 69.7 (0.8) & \textbf{74.1} (0.9) \\ 
        A $\rightarrow$ Bo & 94.3 (0.4) & \textbf{92.5} (1.1) & 90.2 (1.2) & T $\rightarrow$ G & 83.6 (0.7) & 79.3 (0.5) & \textbf{83.6} (0.3) \\ 
        A $\rightarrow$ C & 95.0 (0.2) & 91.8 (0.5) & \textbf{92.1} (0.5) & T $\rightarrow$ S & 72.3 (0.5) & 69.6 (0.1) & \textbf{70.7} (0.6) \\ 
        A $\rightarrow$ M & 85.8 (0.5) & 81.3 (0.6) & \textbf{83.3} (0.5) & T $\rightarrow$ Te & 77.8 (0.1) & 69.4 (0.8) & \textbf{76.8} (0.0) \\ 
        \midrule
        B $\rightarrow$ A & 93.4 (0.3) & 93.3 (0.2) & \textbf{93.4} (0.4) & F $\rightarrow$ T & 79.9 (0.1) & \textbf{69.9} (0.2) & 65.4 (0.8) \\ 
        B $\rightarrow$ Bo & 94.7 (0.7) & \textbf{93.8} (0.3) & 92.2 (0.1) & F $\rightarrow$ G & 82.3 (0.1) & 54.3 (23.4) & \textbf{78.8} (2.5) \\ 
        B $\rightarrow$ C & 94.7 (0.8) & \textbf{93.4} (0.1) & 92.1 (0.3) & F $\rightarrow$ S & 72.1 (0.2) & 64.6 (1.8) & \textbf{65.3} (1.6) \\ 
        B $\rightarrow$ M & 85.3 (0.2) & 81.3 (0.7) & \textbf{82.8} (0.2) & F $\rightarrow$ Te & 78.3 (0.6) & 64.6 (0.7) & \textbf{72.5} (0.2) \\
        \midrule
        Bo $\rightarrow$ A & 94.6 (0.3) & \textbf{91.6} (0.5) & 91.3 (0.2) & G $\rightarrow$ T & 79.9 (0.4) & \textbf{75.9} (0.3) & 75.8 (0.6) \\ 
        Bo $\rightarrow$ B & 94.8 (0.2) & \textbf{92.9} (0.6) & 90.9 (0.2) & G $\rightarrow$ F & 76.7 (0.1) & 69.9 (0.2) & \textbf{73.5} (0.2) \\ 
        Bo $\rightarrow$ C & 94.3 (0.2) & 89.8 (0.1) & \textbf{90.3} (0.4) & G $\rightarrow$ S & 73.1 (0.0) & \textbf{69.4} (0.1) & 68.0 (1.8) \\ 
        Bo $\rightarrow$ M & 85.5 (0.9) & \textbf{84.6} (0.7) & 80.1 (1.2) & G $\rightarrow$ Te & 78.1 (0.6) & 69.9 (0.3) & \textbf{73.5} (0.6) \\ 
        \midrule
        C $\rightarrow$ A & 93.4 (0.4) & 92.3 (0.3) & \textbf{92.5} (0.6) & S $\rightarrow$ T & 79.5 (0.3) & 74.4 (1.7) & \textbf{76.8} (0.1) \\ 
        C $\rightarrow$ B & 95.0 (0.6) & \textbf{94.1} (0.1) & 92.1 (0.2) & S $\rightarrow$ F & 77.7 (0.2) & \textbf{73.1} (0.0) & 72.4 (0.5) \\ 
        C $\rightarrow$ Bo & 93.9 (0.8) & \textbf{91.3} (0.5) & 89.0 (0.1) & S $\rightarrow$ G & 83.4 (0.2) & \textbf{78.2} (0.5) & 76.3 (0.9) \\ 
        C $\rightarrow$ M & 85.8 (0.1) & \textbf{81.5} (0.7) & 79.7 (1.2) & S $\rightarrow$ Te & 78.5 (0.0) & 66.7 (0.2) & \textbf{74.8} (0.3) \\ 
        \midrule
        M $\rightarrow$ A & 94.2 (0.7) & 89.1 (1.4) & \textbf{90.1} (0.5) & Te $\rightarrow$ T & 79.8 (0.3) & 71.4 (0.0) & \textbf{76.5} (0.4) \\ 
        M $\rightarrow$ B & 95.3 (0.5) & 81.0 (16.1) & \textbf{89.9} (1.2) & Te $\rightarrow$ F & 77.9 (0.1) & 69.9 (0.5) & \textbf{74.3} (0.5) \\ 
        M $\rightarrow$ Bo & 94.1 (0.4) & 80.5 (18.6) & \textbf{91.5} (0.0) & Te $\rightarrow$ G & 82.5 (0.1) & 75.6 (1.6) & \textbf{82.0} (0.6) \\ 
        M $\rightarrow$ C & 94.3 (0.5) & \textbf{90.5} (0.0) & 89.7 (0.3) & Te $\rightarrow$ S & 72.2 (0.0) & 68.0 (0.4) & \textbf{71.3} (0.5) \\ 
        \bottomrule
    \end{tabular}
    }
    \caption{Comparison of \mlm~and \mmd by target domain classification accuracy on the Amazon Product Review and MNLI datasets. Each row represents a Source$\to$ Target pair. On average, \mlm~is competitive with \mmd, often outperforming it.
    We use the T5v1.1 base model, and (IA)$^3$ as a PEFT method.
    The highest values between \mlm~and \mmd~have been marked in bold. 
    }
    \label{tab:results-combined-no-src-only}
\end{table*}

\begin{figure}
\begin{center}
\centerline{
\includegraphics[width=\columnwidth]{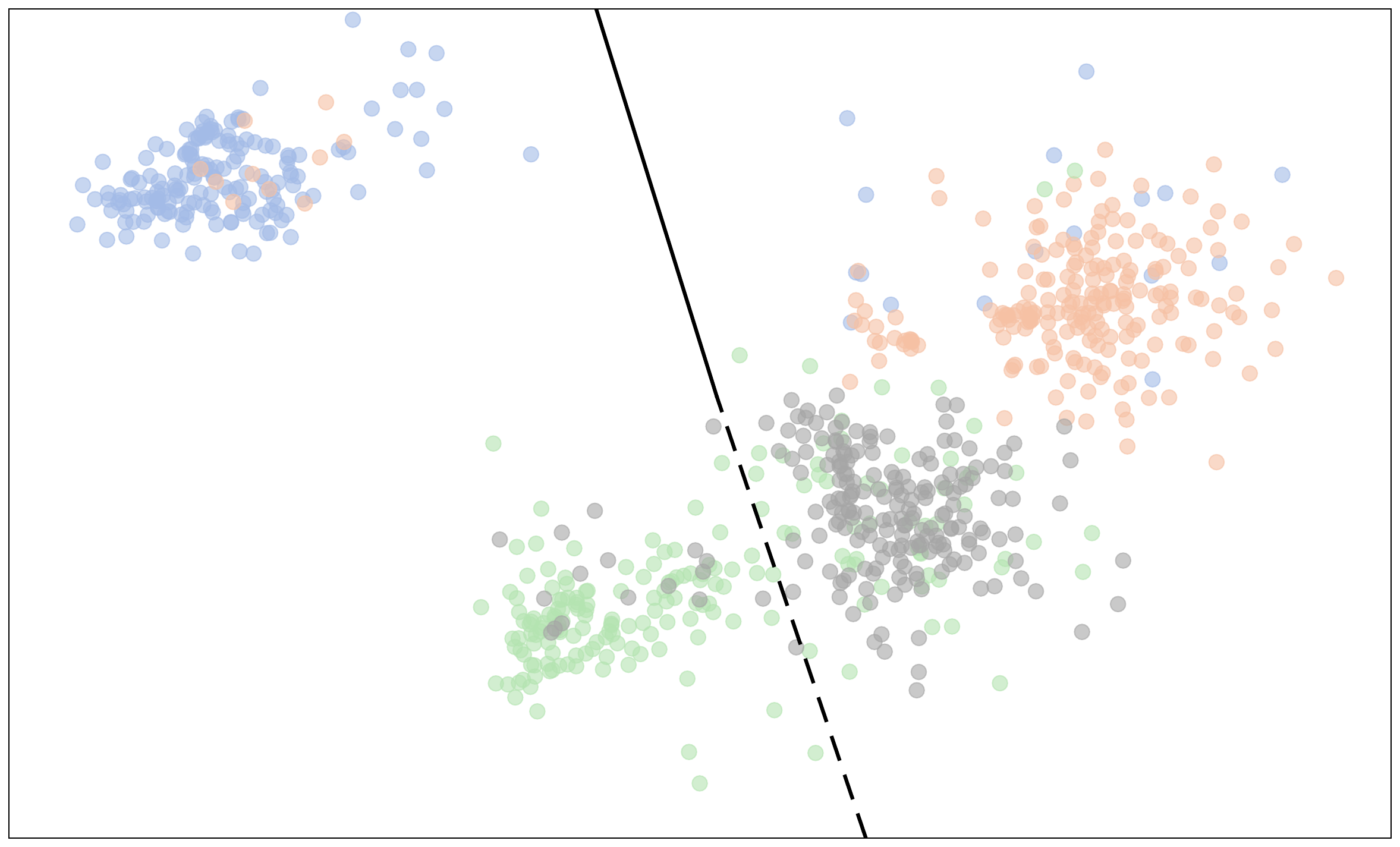}
}
\centerline{
\includegraphics[width=\columnwidth]{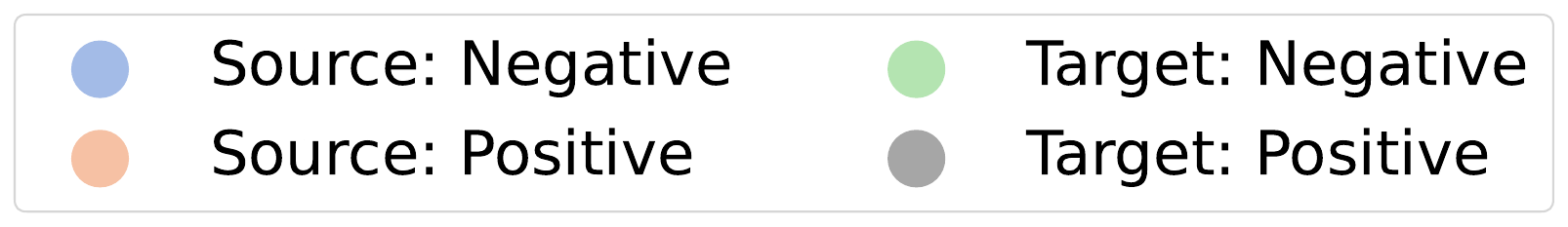}
}
\caption{UMap visualizations of sentence embeddings from the Apparel $\to$ Movies data pair, using the T5v1.1 base model and (IA)$^3$ PEFT method.
Despite not promoting domain-invariance, \mlm may be learning sentence embeddings that are separable by class labels, regardless of the domain of these sentences. 
The classification hyperplane for the source domain has been imagined as a solid line for illustration purposes, and its extension to the target domain is shown as a dashed line.
}
\label{fig:mlm-understanding}
\end{center}
\end{figure}

\begin{table}[!ht]
    \centering
    \resizebox{0.4\textwidth}{!}{
    \begin{tabular}{ll}
    \toprule
    \textbf{Method} & \textbf{Accuracy} \\ 
    \midrule
        \mlm & \textbf{83.3} (0.9) \\
        \mmd & 81.3 (0.6) \\ 
        DANN & 52.3 (1.7) \\
        CORAL & 80.9 (0.4) \\
        Task Vectors & 48.0 (0.7)\\
        Task Vectors (fine-tuning) & 69.0 (0.4) \\
    \bottomrule
    \end{tabular}
    }
\caption{Comparison of \mlm~with more baselines, using the T5v1.1 base model and (IA)$^3$ PEFT method on the Apparel$\to$Movies pair from the Amazon review dataset. For task vectors, we include versions with (IA)$^3$ as well as full fine-tuning.
\mlm~outperforms all baselines.
}
\label{tab:more-baselines}
\end{table}

\paragraph{Comparison with other Model Centric Approaches} In addition to the MMD based method of~\citet{malik2023udapter}, we also compare~\mlm with other methods that promote domain invariance: 1) DANN~\citep{ganin2016domain}, which is the most widely used UDA method in NLP~\citep{ramponi2020neural}, but has been shown to be highly unstable; 2) CORAL~\citep{sun2017correlation}, which minimizes second order statistics of the data embeddings. Additionally, with an emerging class of weight interpolation based methods, we make a comparison with task vector arithmetic~\cite{ilharco2022editing}. The use of task vectors with PEFT methods beyond LoRA~\citep{hulora} has been unexplored in the literature, and we find that the method does not work with IA3. With fully fine-tuned models, the method improves in performance, but is still weaker than~\mlm. 

\paragraph{\mlm may learn representations that generalize across domains} To better understand the improved UDA performance, we visualize the sentence embeddings learned by \mlm in Figure~\ref{fig:mlm-understanding}. 
Using UMap~\cite{mcinnes2018umap}, the figure visualizes embeddings for the Apparel$\to$Movies domain pair from the Amazon Product Review dataset.
We see that \mlm~ learns sentence embeddings that generalize across domains. For illustration, we draw a black line that cuts across both source and target domains. Note that the solid line suggests that there exists a classification hyperplane learned on the source labeled data (in \textcolor{blue}{blue} and \textcolor{teal}{green}). The same classifier can be potentially used to separate target data (in \textcolor{gray}{gray} and \textcolor{orange}{orange}). The visualization suggests that \mlm achieves competitive UDA results without having to explicitly promote domain-invariant representations. 

\section{\mlm across Model Architectures and Scales}
\label{sec:appendix-other-models}

We evaluate the performance of \mlm~over T5v1.1 XL and the instruction tuned T0 (3B)~\citep{sanhmultitask} in Table~\ref{tab:ins-tuned-models}.

\begin{table}[!ht]
    \centering
    \resizebox{0.5\textwidth}{!}{
    \begin{tabular}{lllll}
    \toprule
        \textbf{Model} & \textbf{\supervised} & \textbf{\mmd} & \textbf{\mlm} \\ 
        \midrule
        T5 v1.1 Base & 85.8 (0.5) & 78.6 (1.3) & \textbf{83.3} (0.5) \\ 
        T5 v1.1 XL & 93.0 (0.5) & 65.2 (9.5) &\textbf{92.0} (1.5)  \\ 
        T0 3B & 92.2 (0.7) & 51.8 (0.8) & \textbf{93.8} (0.4)  \\     \bottomrule
    \end{tabular}
    }
\caption{The performance gap between \mlm~and \mmd~increases with larger models, from T5v1.1 Base (60M parameters) to T5v1.1 XL (3B parameters), and further increases with instruction tuning (T0 3B).
}
\label{tab:ins-tuned-models}
\end{table}

\section{PEFT Frameworks}
\label{sec:appendix-peft-frameworks}

The framework proposed in Section~\ref{sec:methodology} is general and can be applied to fine-tune all model parameters. Additionally, our \mlm framework is compatible with the parameter-efficient fine-tuning approach. The PEFT approach is desirable because it adds only a small amount of learnable parameters $\phi$ to a pre-trained language model $\theta$, 
and fine-tunes only $\phi$ to perform prediction while keeping the other model parameters $\theta$ frozen. 
We use two instantiations in our implementations: Adapters \cite{houlsby2019parameter} and (IA)$^3$ \cite{liu2022few}.

\paragraph{(IA)$^3$} is a state of the art PEFT learning method, and uses around a tenth of learnable parameters compared to popular methods like Adapters. (IA)$^3$ works by element-wise multiplication (i.e. rescaling) of the model’s activations against a learned vector. In this case, the set of learnable parameters $\phi$ is a set of vectors $\{l_\text{v}, l_\text{k}, l_{\text{ff}}\}$ applied to each attention mechanism and feed-forward layer as,

\newcommand*{\elementwiseprod}{%
    \mathbin{%
        \ooalign{$\circledcirc$\cr\hidewidth$\bullet$\hidewidth}%
    }%
}
\begin{align*}
    h &= \sigma \left( \frac{Q (l_\text{k} \elementwiseprod K^T)}{\sqrt{d_k}} \right) (l_\text{v} \elementwiseprod V) \\
    h &= (l_\text{ff} \elementwiseprod \gamma (W_1 x) W_2)
\end{align*}
Here, $K$, $Q$ and $V$ are the key, query and value representations used in an attention block, and  $W_1$ and $W_2$ are the weights in the feed-forward layer following an attention block. $l_\text{k} \in \mathbb{R}^{d_k}$, $l_\text{v} \in \mathbb{R}^{d_v}$, $l_\text{ff} \in \mathbb{R}^{d_\text{ff}}$, $\sigma$ is the softmax function while $\gamma$ is any non-linearity. 

Intuitively, each vector $l$ simply learns weights measuring the importance of each feature in an activation of the pre-trained model, for the specific downstream task the model is trained on.

\paragraph{Adapters} are a popularly used and high performing PEFT framework, and \citet{hetowards} have shown equivalence in the operations applied by Adapters, Prefix Tuning \cite{li2021prefix} and LoRA \cite{hulora}. 

Adapters work by adding small learnable modules between transformer layers. Specifically, down and up projections $W_\text{down} \in \mathbb{R}^{d \times r}$ and $W_\text{up} \in \mathbb{R}^{r \times d}$ are learnt such that $\phi = \{W_\text{up}, W_\text{down}\}$. A residual connection and non-linearity $\gamma$ is added at every layer, 
\begin{equation*}
    h = h + \gamma(h W_\text{down}) W_\text{up}
\end{equation*}

Table~\ref{tab:varying-peft} shows \mlm~beats \mmd across different tuning methods. We also note that fine-tuning yields slightly better performance for all UDA methods. 

\begin{table*}[!ht]
    \centering
    \resizebox{0.5\textwidth}{!}{
    \begin{tabular}{llll}
    \toprule

        \textbf{Method} & \textbf{\supervised} & \textbf{\mmd} & \textbf{\mlm} \\ \midrule
        Fine-Tuning & 86.4 (0.4) & 82.4 (1.6) & \textbf{84.4} (0.3) \\ 
        (IA)3 & 85.8 (0.5) & 81.3 (0.6) & \textbf{83.3} (0.5) \\ 
        Adapters & 85.3 (0.5) & 79.1 (0.3) & \textbf{82.7} (0.5) \\
        \bottomrule
    \end{tabular}
    }
    \caption{Performance of \mlm~across different adaptation methods
    with the T5v1.1 base model on the Apparel $\to$ Movies domain pair. \mlm~remains more powerful than \mmd across all methods.
    }
    \label{tab:varying-peft}
\end{table*}

\section{\mlm in a Few-Shot Setup}
\label{sec:appendix-few-shot-learning}

Table~\ref{tab:k-shot} accompanies Figure~\ref{fig:ins-tuned-models} (Section~\ref{sec:results-more-settings}), showing the 256-shot performance of~\mlm and other baselines, across model sizes. Similarly, Table~\ref{tab:performance-across-shots} accompanies Figure~\ref{fig:256-shot-ins-tuned-models}, showing the relative performance of all baselines across varying $k$.

\begin{table*}[!ht]
    \centering
    \resizebox{\textwidth}{!}{
    
    \begin{tabular}{llll}
    \toprule
        \multirow{1}{*}{\textbf{Model}} & \multicolumn{1}{c}{\textbf{\supervised}} & \multicolumn{1}{c}{\textbf{\mmd}} & \multicolumn{1}{c}{\textbf{\mlm}}  \\ 
        \midrule
        T5v1.1 Base & 77.8 (0.4) & 60.9 (1.6) & 73.1 (1.7) \\
        T5v1.1 XL & 84.4 (0.1) & 84.8 (1.5) & 89.9 (1.1) \\ 
        T0 3B & 88.3 (0.5) & 81.8 (1.3) & 93.9 (0.4) \\ 
        \bottomrule
    \end{tabular}
    
    \begin{tabular}{lllll}
    \toprule
        \multirow{1}{*}{\textbf{Model}} & \multicolumn{1}{c}{\textbf{\supervised}} & \multicolumn{1}{c}{\textbf{\mmd}} & \multicolumn{1}{c}{\textbf{Two Phase}} & \multicolumn{1}{c}{\textbf{\mlm}}  \\ 
        \midrule
        T5v1.1 Base & 82.8 (0.6) & 62.5 (0.7) & 79.8 (1.4) & 81.2 (0.7) \\
        T5v1.1 XL & 92.5 (0.4) & 71.7 (7.8) & 84.3 (0.9) & 86.8 (2.2) \\ 
        T0 3B & 91.8 (0.6) & 79.5 (6.7) & 53.5 (0.4) & 92.8 (0.2) \\ 
        \bottomrule
    \end{tabular}
    }
    \caption{Performance of \mlm across different models, in a k-shot learning setup on the Apparel $\to$ Movies domain pair.
    We see \mlm retaining strong performance on the target domain across models. \textbf{Left}: 32-shot. \textbf{Right}: 256-shot.}
    \label{tab:k-shot}
\end{table*}
\begin{table*}[!ht]
    \centering
    \resizebox{0.75\textwidth}{!}{
    \begin{tabular}{lllllllllll}
    \toprule
        \multirow{1}{*}{\textbf{Number of Shots}} & \multicolumn{1}{c}{\textbf{\supervised}} & \multicolumn{1}{c}{\textbf{\mmd}} & \multicolumn{1}{c}{\textbf{Two Phase \mmd}} & \multicolumn{1}{c}{\textbf{\mlm}}  \\ 
        \midrule
        32 & 77.8 (0.4) & 60.9 (1.6) & 59.4 (2.0) & 73.1 (1.7) \\
        128 & 82.5 (0.5) & 75.1 (0.6) & 62.8 (0.6) & 78.8 (1.0) \\ 
        256 & 82.8 (0.6) & 62.5 (0.7) &79.8 (1.4) & 81.2 (0.7) \\
        \bottomrule
    \end{tabular}
    }
    \caption{Performance of \mlm across different number of shots, on the Apparel $\to$ Movies domain pair, using the T5v1.1 base model.
    We see \mlm retaining strong performance on the target domain across shots.}
    \label{tab:performance-across-shots}
\end{table*}

\section{Impact of Target Domain Exposure}
\label{sec:appendix-mask-rate}

The experiments in this section use the T5v1.1 base model on the Apparel$\rightarrow$Movies domain pair of the Amazon reviews dataset. 

Table~\ref{tab:masking-rate} accompanies results from Figure~\ref{fig:mask-rate-impact}, which show the impact of varying masking rates on \mlm.
Using the T5v1.1 base model, we train \mlm using varying random masking rates on the Apparel $\to$ Movies domain pair, and report the mean and standard deviation over three runs. 
With high masking rates, the performance on the source domain is largely maintained, but the performance on the target domain rapidly deteriorates.

\begin{table}[!ht]
    \centering
    \resizebox{0.4\textwidth}{!}{
    \begin{tabular}{lll}
    \toprule
        \multirow{2}{*}{\textbf{Masking Rate}} & \multicolumn{2}{c}{\textbf{Accuracy}}  \\ 
        ~ & \multicolumn{1}{c}{Source} &\multicolumn{1}{c}{Target} \\ 
    \midrule
        \multicolumn{1}{c}{5\%} & 92.8 (0.8) & 78.8 (1.8) \\
        \multicolumn{1}{c}{15\%} & \textbf{93.5} (0.4) & \textbf{83.3} (0.5) \\
        \multicolumn{1}{c}{30\%} & 92.8 (0.6) & 78.8 (1.4) \\ 
        \multicolumn{1}{c}{60\%} & 92.5 (0.9) & 71.0 (3.0) \\ 
        \multicolumn{1}{c}{90\%} & 92.3 (0.5) & 70.4 (1.5) \\ 
    \bottomrule
    \end{tabular}}
\caption{Impact of Masking Rate on \mlm.
We train \mlm using varying random masking rates on the Apparel $\to$ Movies domain pair.
With high masking rates, the performance on the source domain is largely maintained, but the performance on the target domain rapidly deteriorates.
}
\label{tab:masking-rate}
\end{table}

\section{Understanding how~\mlm aids UDA}
\label{sec:appendix-mlm-analysis}

Table~\ref{tab:masking-inference} (accompanies Figure~\ref{fig:info-vs-uninfo-masking}) shows the impact of masking sequences at inference, on classification accuracy. 
Words are selected for masking based on their their ``informativeness'', measured by their PMI to the inference class label. 
The performance of the model is best with the original unmasked sequences, indicating the presence of both informative and uninformative words are essential for strong classification performance.

Table~\ref{tab:few-shot-masking-analysis} accompanies Figure~\ref{fig:info-vs-uninfo-masking} and shows the impact of varying masking strategies on classification performance, in a few-shot setting. We also consider two different few-shot setups: one with access to the full unlabelled datasets in phase 1 pre-training, and another where even the unlabelled data is few-shot. 

To isolate any effects of PEFT methods or pre-training data, we repeat the analysis from Table~\ref{tab:few-shot-masking-analysis} in Table~\ref{tab:informative-vs-uninformative-sanity} with fine-tuning Flan-T5 in a full data setting, and note similar trends. 

\begin{table}[!ht]
    \centering
    \resizebox{0.36\textwidth}{!}{
    \begin{tabular}{lll}
    \toprule
    \multirow{2}{*}{\textbf{Method}} & \multicolumn{2}{c}{\textbf{Accuracy}}  \\ 
        ~ & \multicolumn{1}{c}{Source} &\multicolumn{1}{c}{Target} \\ 
    \midrule
        Original & \textbf{93.5} & \textbf{83.3} \\
        Informative Masking & 88.0 & 78.8 \\ 
        Uninformative Masking & 92.0 & 79.0 \\
    \bottomrule
    \end{tabular}
    }
\caption{Impact of masking at inference. 
We evaluate \mlm on the Apparel $\to$ Movies domain pair, and select words for masking based on their ``informativeness'' to the classification task.
}
\label{tab:masking-inference}
\end{table}

\begin{table}[!ht]
    \centering
    \resizebox{0.43\textwidth}{!}{
    \begin{tabular}{lll}
    \toprule
    \multirow{2}{*}{\textbf{Masking Strategy}} & \multicolumn{2}{c}{\textbf{Accuracy}}  \\ 
        ~ & \multicolumn{1}{c}{Source} &\multicolumn{1}{c}{Target} \\ 
    \midrule
        Random & \textbf{95.8} (0.0) & \textbf{86.8} (0.3) \\
        Informative & 93.9 (0.6) & 85.3 (0.3) \\ 
        Uninformative & 95.0 (0.0) & 84.8 (0.1) \\
    \bottomrule
    \end{tabular}
    }
\caption{Impact of word selection for masking during training, using Flan-T5 base and no PEFT methods.
}
\label{tab:informative-vs-uninformative-sanity}
\end{table}
\begin{table}[!ht]
    \centering
    \resizebox{\columnwidth}{!}{
    \begin{tabular}{llll}
    \toprule
    \multirow{2}{*}{\textbf{Phase 1 Data}} & \multirow{2}{*}{\textbf{Masking Strategy}} & \multicolumn{2}{c}{\textbf{Accuracy}}  \\ 
        ~ & ~ & \multicolumn{1}{c}{Source} &\multicolumn{1}{c}{Target} \\ 
    \midrule
            256 Shot & Random & \textbf{91.0} (0.9) & \textbf{78.1} (2.4) \\ 
        ~ & Informative & 90.4 (0.5) & 76.0 (0.7) \\ 
        ~ & Uninformative & 89.6 (1.2) & 73.5 (1.6) \\ 
        \midrule
        Full Data & Random & 90.1 (0.5) & \textbf{81.2} (0.7) \\ 
        ~ & Informative & \textbf{91.8} (0.5) & 78.0 (0.9) \\ 
        ~ & Uninformative & 89.3 (0.5) & 72.8 (1.1) \\
    \bottomrule
    \end{tabular}
    }
\caption{Impact of word selection for masking, in a 256-shot learning setup. 
We evaluate \mlm on the Apparel $\to$ Movies domain pair, and select words for masking based on their ``informativeness'' to the classification task.
Random masking is most powerful for the target domain, indicating that both semantic and background features are necessary for effective classification on the unlabelled domain. 
However, informative masking is significantly more useful than uninformative masking.
}
\label{tab:few-shot-masking-analysis}
\end{table}

\section{Single Phase~\mlm Training}
\label{sec:single-phase-mlm}

Our proposed approach in Section~\ref{sec:methodology} involves two stages of training, which is more expensive than standard single phase UDA approaches. 
In this section, we propose a single training phase variant to \mlm, and show that it performs similarly to the original method. We use the two phase pipeline in our experiments in the main paper, but note that the single and two phase pipelines are interchangeable. 

We simply replace the two phase training with a joint multi-task objective as follows,
\begin{align*}
    \mathcal{L(\mathcal{D}, \mathcal{D}_\text{src}; \theta)} &= 
    \frac{1}{|D|} \frac{1}{|D_\text{src}|} 
    \sum_{x' \in \mathcal{D}} \sum_{(x, y) \in \mathcal{D}_\text{src}} \\
    & (\lambda \; l(\mathbb{C}(x, y); \theta) \\
    & + (1 - \lambda) \; l(\mathbb{M}(x'); \theta))
\end{align*}
where $l$ is the cross-entropy loss defined in Eq. (\ref{equn:cross-entropy}), and $\mathbb{M}$ and $\mathbb{C}$ are the templates defined in Section~\ref{sec:methodology}.
$\lambda$ is the adaptation factor which gradually changes from 0 to 1 over the course of training. This results in the model being trained almost exclusively on the MLM task early on in training, and the CLS task towards the end of training.

Table~\ref{tab:mlm-single-vs-two-phase} compares the performance of the single phase and two phase variants of \mlm. We also compare with a vanilla joint single phase objective, where $\lambda$ is fixed at 0.5 through training (called Single Phase Vanilla). 
The performance of the single and two phase variants are almost identical, and either can be used interchangeably. In comparison, the vanilla single phase method is significantly weaker on the target domain.

\begin{table}[!ht]
    \centering
    \resizebox{0.4\textwidth}{!}{
    \begin{tabular}{lll}
    \toprule
    \multirow{2}{*}{\textbf{Method}} & \multicolumn{2}{c}{\textbf{Accuracy}}  \\ 
        ~ & \multicolumn{1}{c}{Source} &\multicolumn{1}{c}{Target} \\ 
    \midrule
        Two Phase & 93.7 (0.3) & \textbf{83.3} (0.9) \\ 
        Singe Phase & 93.5 (0.4) & \textbf{83.3} (0.5) \\ 
        Singe Phase Vanilla & 93.6 (0.1) & 75.0 (5.7) \\
    \bottomrule
    \end{tabular}
    }
\caption{Comparison of single and two-phase variants of \mlm, on the Apparel $\to$ Movies domain pair. The single and two phase variants are almost identical in performance.}
\label{tab:mlm-single-vs-two-phase}
\end{table}

\section{Instability of Domain Invariance Methods for UDA}
\label{sec:mmd-variations}

The Maximum Mean Discrepancy (MMD) \cite{gretton2012kernel} measures the difference between first order moments of variables in a Reproducing Kernel Hilbert Space \cite{aronszajn1950theory}. 
Multiple lines of work have shown that minimizing divergence measures like MMD, when combined with auxiliary task-specific loss functions, results in training instabilities and vanishing gradients~\cite{ramesh2021domain, han2019unsupervised}.

\begin{figure}[ht]
\begin{center}
\centerline{\includegraphics[width=0.4\textwidth]{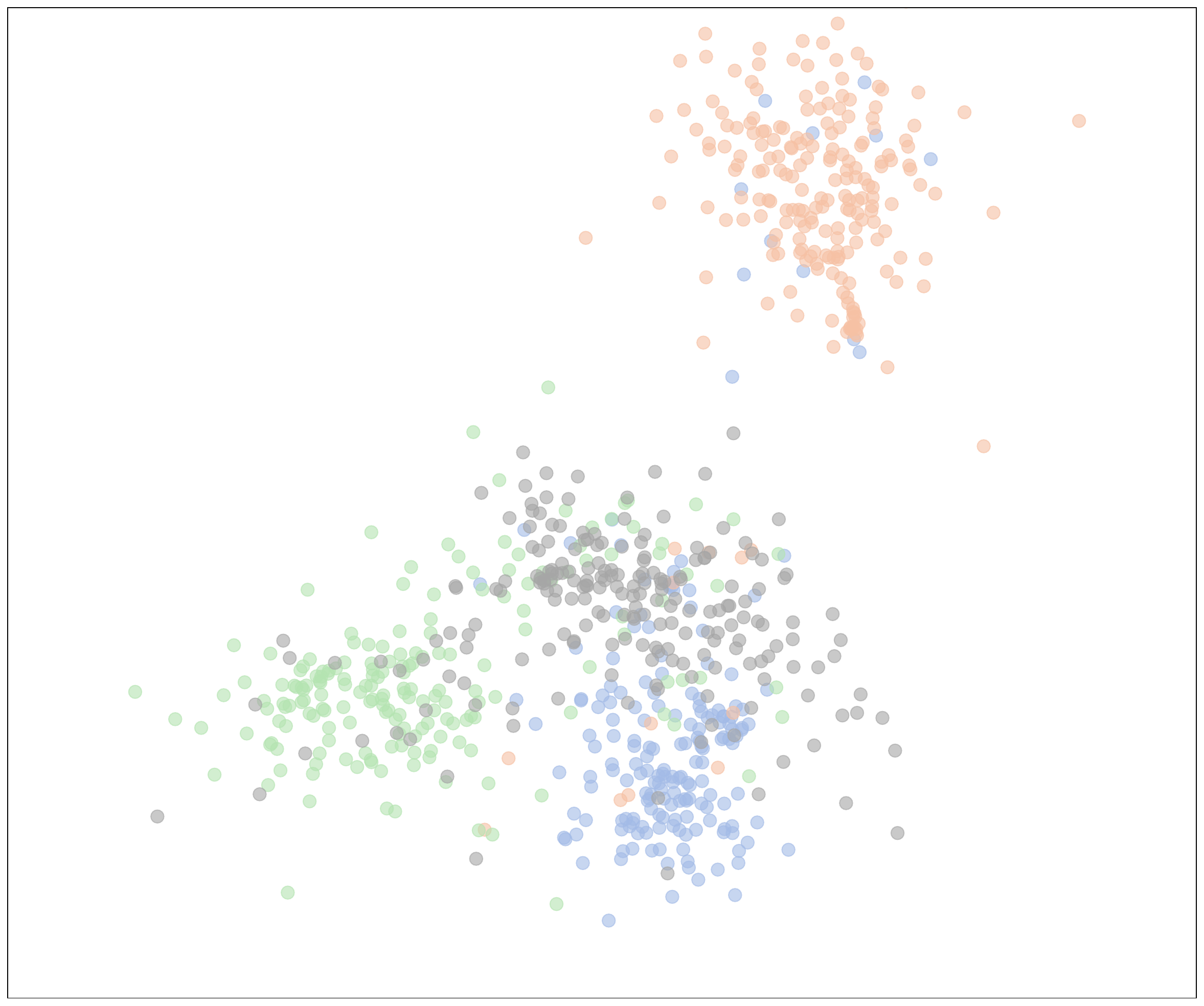}}
\centerline{
\includegraphics[width=0.5\textwidth]{figures/images/apparel_movies/legend.pdf}
}
\caption{UMap visualizations of sentence embeddings from the Apparel $\to$ Movies data pair, using the T5v1.1 base model and (IA)$^3$ PEFT method. Training with \mmd risks stability issues, and all embeddings from the target domain can be mapped to the closest source class cluster. This results in poor classification performance on the target domain.}
\label{fig:mmd-unstable}
\end{center}
\end{figure}

We also note that as minimizing MMD does not use any label information, there is a possibility for embeddings of the target domain to be aligned with the closest source domain class cluster. For example, Figure~\ref{fig:mmd-unstable} shows us a setting where both classes of the target domain (shown in \textcolor{teal}{green} and \textcolor{gray}{gray}) are mapped to the cluster of negative class source embeddings (shown in \textcolor{blue}{blue}). 

We compare variants of the \mmd method in Table~\ref{tab:mmd-variants} and show that the loss is sensitive to small changes in the loss design. Specifically we compare the \mmd method used in the main paper with: 
\begin{itemize}
    \item MMD over Logits: Measures the MMD between the logits of source and target domains, instead of using intermediate model outputs.
    \item Fixed Weight MMD: Instead of the multi-task loss for the MMD reduction and classification tasks, we use fixed weights for both tasks\footnote{For the weighted loss, $\mathcal{L}_\text{CLS} + 3 \; \mathcal{L}_\text{MMD}$ was found to be the best performing.}.
    \item Two Phase MMD: The first training phase is used to minimize MMD between source and target embeddings, while the second phase is used to train the model for classification on the source domain. 
\end{itemize}
\mlm remains more powerful than all variants.

\begin{table}[!ht]
    \centering
    \resizebox{0.9\columnwidth}{!}{
    \begin{tabular}{lll}
    \toprule
    \multirow{2}{*}{\textbf{Method}} & \multicolumn{2}{c}{\textbf{Accuracy}}  \\ 
        ~ & \multicolumn{1}{c}{Source} &\multicolumn{1}{c}{Target} \\ 
    \midrule
        \mlm & 93.7 (0.3) & \textbf{83.3} (0.9) \\ 
        \mmd & \textbf{94.7} (0.3) & 81.3 (0.6) \\ 
        \mmd over Logits & 95.0 (0.2) & 81.3 (0.7) \\ 
        Fixed Weight MMD & 93.4 (0.2) & 78.6 (1.3) \\
        Two Phase \mmd & 90.1 (0.1) & 68.7 (2.0) \\ 
    \bottomrule
    \end{tabular}
    }
\caption{Comparison of variants of minimizing MMD, on the Apparel $\to$ Movies domain pair. \mlm remains more powerful than all variants.}
\label{tab:mmd-variants}
\end{table}

\end{document}